\documentclass[runningheads]{llncs}
\usepackage[T1]{fontenc}
\usepackage{url}
\usepackage{xcolor}
\usepackage{caption}
\usepackage{subcaption}

\usepackage{graphicx}
\usepackage{tabularx}
\usepackage{multirow}
\usepackage{graphicx}
\begin{document}
\title{Human Fall Detection using Transfer Learning-based 3D CNN}
\author{Ekram Alam\inst{1,3}\orcidID{0000-0003-1599-1401} \and
Abu Sufian\inst{2}\orcidID{0000-0003-2035-2938} \and
Paramartha Dutta\inst{3} \orcidID{0000-0003-3946-2440} \and Marco Leo \inst{4} \orcidID{0000-0001-5636-6130} }
\authorrunning{Ekram Alam et al.}

 % abbreviated author list (for running head)


%

\institute{Department of Computer Science, Gour Mahavidyalaya,Old Malda, India 
	\email{ealam@ieee.org}
	%\texttt{http://users/\homedir iekeland/web/welcome.html}
	\and
	Department of Computer Science, University of Gour Banga, English Bazar, India 
	\and Department of Computer and System Sciences, Visva-Bharati University, Santiniketan, India
	\and National Research Council of Italy, Institute of Applied Sciences and Intelligent Systems, 
	Lecce, Italy }

\maketitle              % typeset the header of the contribution
\begin{abstract}
Unintentional or accidental falls are one of the significant health issues in senior persons. The population of senior persons is increasing steadily. So, there is a need for an automated fall detection monitoring system. This paper introduces a vision-based fall detection system using a pre-trained 3D CNN. Unlike 2D CNN, 3D CNN extracts not only spatial but also temporal features. Proposed model leverages the original learned weights of a 3D CNN model pre-trained on the Sports1M dataset to extract the spatio-temporal features. Only the SVM classifier was trained, which saves the time required to train the 3D CNN. Stratified shuffle five split cross-validation has been used to split the dataset into training and testing data. Extracted features from the proposed 3D CNN model were fed to an SVM  classifier to classify the activity as fall or ADL. Two datasets, GMDCSA and CAUCAFall, were utilized to conduct the experiment. The source code for this work can be accessed via the following link: \url{https://github.com/ekramalam/HFD_3DCNN}.

\end{abstract}
\section{Introduction}
According to a report by the ``World Health Organization (WHO)'' \cite{whoAgeingHealth}, the global population of individuals aged 60 years or above is projected to increase from 1 billion in 2020 to 1.4 billion in 2030 and is expected to rise further to 2.1 billion by 2050. The steady rise in the global population of senior citizens poses a significant challenge in caring for seniors. Accidental or unintentional falls represent a significant health concern among senior individuals, often resulting in severe injuries, permanent damage, escalated healthcare expenses, and, in some cases, even death if timely medical assistance is delayed \cite{martinez2019up}. An automated monitoring system can be exceedingly beneficial for detecting human falls among senior citizens or individuals with illnesses who live alone.

Among various systems designed for detecting falls, vision-based and wearable sensor-based fall detection systems are two major approaches. Wearable sensors, while promising in many aspects, present several disadvantages for detecting human falls. Wearable sensors can be annoying and uncomfortable to wear continuously, leading to hesitance in embracing this. Prolonged use may result in side effects like skin irritation or discomfort, potentially causing ill effects on health, particularly for those with sensitive skin or pre-existing conditions. Additionally, frequent battery charging can be a hassle for seniors, making these devices less user-friendly and less practical for those with limited mobility or cognitive challenges.

In the vision-based approach, there is no need to attach the sensor to the body. Vision sensors usually do not require frequent battery charging, further enhancing convenience and long-term viability. Vision sensors are becoming increasingly prevalent in various settings, including homes, hospitals, nursing homes, industries, and public spaces, generating vast amounts of continuous data. This widespread availability and self-sustaining nature make vision sensor-based methods highly promising for human fall detection. Vision sensor-based approaches are also cost-effective \cite{paulauskaite2023geriatric}.

Machine Learning (ML) based techniques, especially data-driven approaches like Deep Learning (DL), work very well to process visual data. Traditional 2D Convolutional Neural Networks (CNNs) \cite{alam2021leveraging,ghosh2020fundamental} is a popular DL technique in computer vision. 2D CNN works frame by frame. It considers only one frame at a time. One frame provides only spatial information. 2D CNN works well in image data, but 2D CNN is not so good for video data. Though spatial information is important to detect falls, temporal information is also required for a better understanding of fall patterns. The intricate nature of human falls requires models that can effectively capture both temporal and spatial features. To extract the temporal and spatial features (spatio-temporal), 3D CNN \cite{maturana2015voxnet} is a better option \cite{liu2021dynamic}. 3D CNNs have proven to be a superior approach \cite{tran2015learning,vrskova2022human,alanazi2022human} for human fall detection. 

3D CNN-based models demonstrate remarkable results, but like other DL techniques, they also necessitate extensive datasets, extended training time, and substantial computing resources. These issues can be addressed by using transfer learning \cite{sufian2021deep}. We used a 3D CNN model \cite{tran2015learning} pre-trained on the Sports-1M dataset \cite{karpathy2014large}. A linear ``Support Vector Machine (SVM)'' \cite{noble2006support,cervantes2020comprehensive} classifier was employed to classify an activity as a ``fall'' or an ``activity of daily living (ADL)''.

The key highlights of this research are outlined below.
\begin{itemize}
	\item \textbf{3D-CNN:} 3D-CNN gives good results because it extracts spatio-temporal features. We have used 3D CNN  in this work.
	\item \textbf{Transfer Learning:} Transfer learning has been used in this work. This work uses a modified pre-trained C3D \cite{tran2015learning} model.
	\item \textbf{Reduced Training Time:} We used the original weights of C3D in our modified C3D model. Only the SVM classifier was trained. So, training time reduced significantly. 
	\item \textbf{Data Adaptability:} One of the essential features of deep learning models like 3D CNN is data generalization. A pre-trained model on Sports-1M was utilized for this work.
	\item \textbf{Tested on recent complex datasets:} We tested our work on two recent complex datasets \cite{alam2023realtime,guerrero2022dataset}.
	\item \textbf{Good Results:} Even though we used the original weight of the C3D model without any training on the fall datasets \cite{alam2023realtime,guerrero2022dataset}, the outcomes are promising as depicted in Table \ref{T_Res_GMDCSA}, and Table \ref{T_ResultCauca}.
\end{itemize}

The remaining sections of this manuscript are outlined as follows. Section \ref{S_LitReview} furnishes a concise review of relevant literature. Section \ref{S_Mat&Method} outlines the datasets and the method employed in this work. The experiment's outcome is detailed and analyzed in Section \ref{S_Result}. Finally, Section \ref{S_Conclusion} encapsulates the findings of this paper and underscores potential avenues for future research.

\section{Literature Review}
\label{S_LitReview}

The detection of human falls can be achieved through the utilization of wearable sensors or vision sensors. Many fall detection systems have been implemented using wearable sensors \cite{mekruksavanich2023deep,seenath2023conformer,andrade2022human,tahir2022iot}, but vision sensor based systems are more suitable for this task \cite{tran2015learning,vrskova2022human,alanazi2022human}. Brief descriptions of recent and notable fall detection systems utilizing vision sensors and deep learning are provided below.

Rezaee et al. \cite{rezaee2022intelligent}, and Arun et al. \cite{arun2022video} used CNN to detect human falls. Rezaee et al. \cite{rezaee2022intelligent}  used thermal datasets on the modified version of ShuffleNet \cite{zhang2018shufflenet}. Inturi et al. \cite{inturi2023novel} introduced a fall detection system that utilizes pose estimation \cite{alam2023realtime}.
A pre-trained AlphaPose \cite{fang2022alphapose} model was leveraged to estimate the pose (joint key points). CNN and LSTM were employed to process the key points and to find the results.
Saurav et al. \cite{saurav2022vision} introduced a human fall detection technique using many variants models consisting of CNN \& ConvLSTM, CNN \& LSTM, and 3D CNNs.
Patel et al. \cite{patel2022indoor} proposed a human fall detection using ``Long-term Recurrent Convolutional Network (LRCN)'', which is a hybrid version of LSTM and CNN.
%
%Xia et al. \cite{xiao2023towards} presented a work for fall detection using unsupervised domain adaptation (UDA). RGB and depth datasets were used as inputs. 

Fei et al. \cite{fei2023flow} introduced a two-streams model for fall detection. One stream used optical flow, and the other one used pose estimation. Graph Convolutional Networks (GCN) \cite{zhang2019graph} and CNN were used for pose and optical flow, respectively. 
Egawa et al. \cite{egawa2023dynamic} introduced a fall detection system utilizing a spatial-temporal convolutional neural network with attention mechanisms based on graphs. Amsaprabhaa et al. \cite{amsaprabhaa2023multimodal} introduced a multi-modal skeletal gait feature fusion-based human fall detection method. Two types of features were extracted using 1D-CNN and spatio-temporal graph convolution network, which were finally combined to get the results.

Chen et al. \cite{chen2022video} and Osigbesan et al. \cite{osigbesan2022vision} introduced a human fall detection system using pose estimation. In Chen et al. \cite{chen2022video}, at first, the 2D pose was estimated, which was transformed into the 3D pose. This 3D pose data was provided as input to the fall detection network to get the final outputs. Their fall detection network is a fully convolutional architecture with residual blocks.
Osigbesan et al. \cite{osigbesan2022vision} introduced a fall detection method for aviation maintenance personnel. They used 3D-CNN and LSTM to process the pose data.

Li et al. \cite{li2023future} presented a human fall detection network using future frame prediction.
Wu et al. \cite{wu2023video} employed GAN (generative adversarial network) \cite{creswell2018generative} to detect human falls.
Zi et al. \cite{zi2023detecting} introduced a fall detection system, especially for low-lighting environments. Dual illumination, YOLOv7, and Deep SORT techniques were used to process the data.

Leal et al. \cite{leal2023detection} presented a human fall detection technique by combining a 3D-CNN and an RNN. Ha et al. \cite{ha2023fall} introduced a human fall detection model using 3D-CNN and a mixture of experts. 
Alanzi et al. \cite{alanazi2023robust} presented a human fall detection method utilizing four branch-3D-CNN (4S-3D-CNN). They divided the input video into 32-frame groups. Segmentation was done to extract humans from these frames. Segmented frames were converted to four fused images using image fusion, which were finally passed to 4S-3DCNN to get the results.

Many fall detection systems have been implemented using 2D CNNs, with only a few utilizing 3D CNNs. 3D CNNs are known to perform better with visual data, but they often come with higher training times. To address this challenge, we adopted a strategy of leveraging a pre-trained 3D CNN. This approach allows our system to extract spatio-temporal features and save training time.

\section{Materials and Method}
\label{S_Mat&Method}
This section is divided into three subsections. Subsection \ref{SS_Dataset} discusses the two datasets ``GMDCSA'' and ``CAUCAFall'' briefly. Subsection \ref{SS_3DNN} provides a brief description of 3D-CNN. Subsection \ref{SS_Method} describes the methodology of this work. 
\subsection{Dataset}
\label{SS_Dataset}

We utilized two datasets, ``GMDCSA'' \cite{alam2023realtime} and ``CAUCAFall'' \cite{guerrero2022dataset}, to evaluate our experiment. The GMDCSA dataset comprises 16 fall videos and 16 non-fall (ADL) videos. The fall video's length exhibits minimum, maximum, mean, mode, and median values of 4 seconds, 6 seconds, 5.06 seconds, 6 seconds, and 5 seconds, respectively. Similarly, the ADL video's length showcases minimum, maximum, mean, mode, and median values of 3 seconds, 12 seconds, 6.5 seconds, 6 seconds, and 6 seconds, respectively. The frames per second (fps) value for GMDCSA is 30. The GMDCSA was generated by performing ADL and fall activities by a single individual. The primary objective of creating this dataset is to detect false positives. Many ADL activity in this dataset involves sleeping activity, which is very similar to fall activity and difficult to identify as ADL by a model.

\begin{table}[h]
	\footnotesize
	\caption{A brief description of the GMDCSA and the CAUCAFall datasets}
	\label{T_Datasets}
	\begin{tabularx}{\linewidth}{|p{1.9cm}|p{1cm}|X|X|X|X|X|p{1cm}|p{1cm}|p{1cm}|p{1cm}|p{1cm}|}
		\hline
		\textbf{Dataset} & \textbf{Type} & \multicolumn{5}{c|}{\textbf{Length (in Seconds)}} & \textbf{fps} & \textbf{NFl} & \textbf{NAd} & \textbf{NSb} & \textbf{NCm} \\
		\cline{3-7}
		& & \textbf{Min} & \textbf{Max} & \textbf{Mean} & \textbf{Mode} & \textbf{Median} & & & & & \\
		\hline
		GMDCSA & Fall & 4 & 6   & 5.06   & 6   & 5   & 30 & 16 & 16 & 1 & 1 \\
		\hline
		GMDCSA & ADL & 3   & 12   & 6.5   & 6   & 6   & 30 & 16 & 16 & 1 & 1 \\
		\hline
		CAUCAFall & Fall & 5   & 13   & 8.18   & 9 \& 10   & 10   & 20 & 50 & 50 & 10 & 1 \\
		\hline
		CAUCAFall & ADL & 4   & 13   & 9.14   & 10   & 10   & 20 & 50 & 50 & 10 & 1 \\
		\hline
	\end{tabularx}
\end{table}
The CAUCAFall dataset, introduced by Guerrero et al. \cite{guerrero2022dataset}, encompasses fall and ADL activities conducted by ten distinct subjects. This dataset incorporates variations in gender (five males and five females), weights, heights, ages, and subject outfits. The dataset was created in varying lighting conditions. Some of the video sequences involve occlusions. Every subject performs five falls and five non-fall activities, which makes a total of 100 video sequences, 50 ``fall'' and 50 ``non-fall'' video sequences. The minimum, maximum, mean, mode, and median values of the length of the fall video sequences of CAUCAFall datasets are 5 seconds,13 seconds, 8.18 seconds, 9 \& 10 seconds, and 10 seconds, respectively. The minimum, maximum, mean, mode and median values of the length of the ADL video sequences of CAUCAFall datasets are 4 seconds,13 seconds, 9.14 seconds, 10 seconds, and 10 seconds respectively.

Table \ref{T_Datasets} provides a brief overview of both datasets, where NFl, NAd, NSb, and NCm represent the number of fall video sequences, the number of ADL video sequences, the number of subjects, and the number of cameras employed, respectively.

\subsection{3D CNN}
\label{SS_3DNN}

\begin{figure}[!htb]
	
	\centering
	
	\includegraphics[scale=0.65]{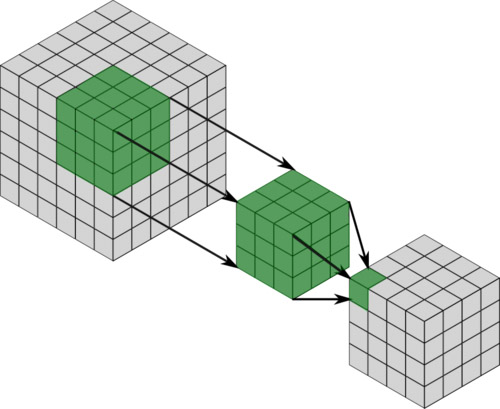}
	\vspace{-5pt}
	\caption{3D Convolution Mechanism}
	
	\label{F_3DConv}
	
\end{figure}

The filter size in 3D convolution and pooling consists of three components, namely ``d x h x w,'' where `h' and 'w' are analogous to the dimensions used in 2D CNNs, representing height and width. The additional component `d' signifies the depth, indicating the number of frames or images. The output of the 3D convolution is not 2D data. Instead, it generates a 3D cuboid output as shown in Figure \ref{F_3DConv}. Similar to 3D convolution, 3D pooling works. 3D Pooling layer downsamples input 3D data.\begin{figure}[!htb]
	
	\centering
	
	\includegraphics[scale=0.6]{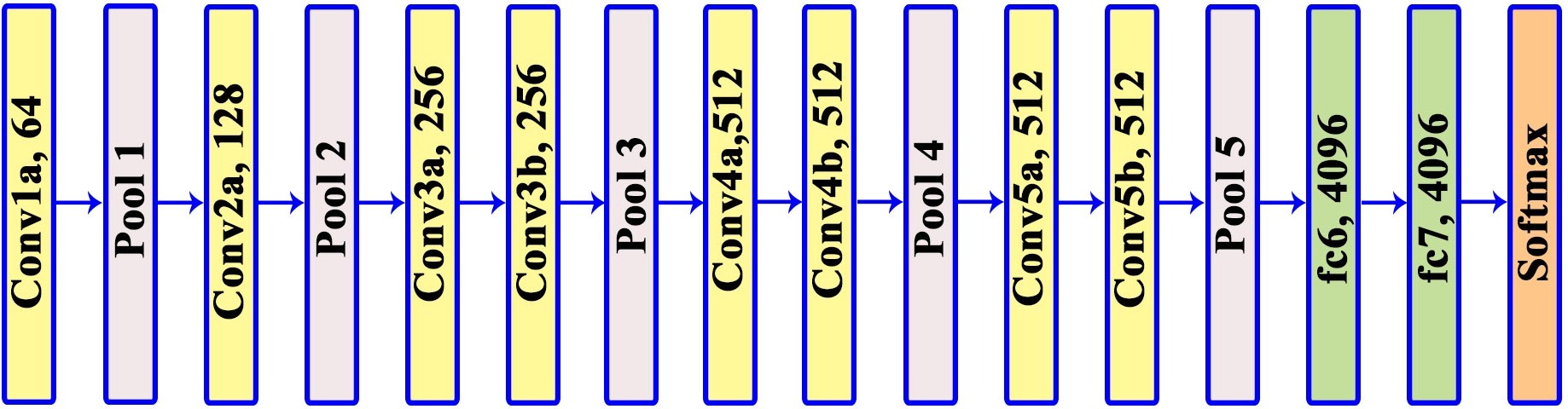}
	\vspace{-5pt}
	\caption{Architecture of the C3D model}
	
	\label{F_C3D}
	
\end{figure} 3D CNNs use 3D convolution and 3D pooling to process the 3D data (video) as 3D volumes, allowing them to comprehend motion dynamics by extracting spatio-temporal features.

Tran et al. \cite{tran2015learning} presented a 3D-CNN with eight convolutions layers (Conv1a, Conv2a, Conv3a, Conv3b, Conv4a, Conv4b, Conv5a, Conv5b), five pooling layers (Pool1, Pool2, Pool3, Pool4, Pool5), two fully connected layers (fc6, fc7), and a Softmax classifier, illustrated in Figure \ref{F_C3D}. They named their 3D-CNN model as C3D. The numbers of filters for the convolution layers Conv1a, Conv2a, Conv3a, Conv3b, Conv4a, Conv4b, Conv5a, and Conv5b were 64, 128, 256, 256, 512, 512, 512, 512 respectively. The size of each kernel for the ``convolutional layer'' was 3 x 3 x 3. The kernel size for the first ``pooling layer'' (Pool1) was 1 x 2 x 2, and for the rest of the ``pooling layers'' it was 2 x 2 x 2. We used the modified version of this C3D model in this work.

\subsection{Method}
\label{SS_Method}
\begin{figure}[!htb]
	\centering % <-- added
	
	\includegraphics[width=.8\linewidth]{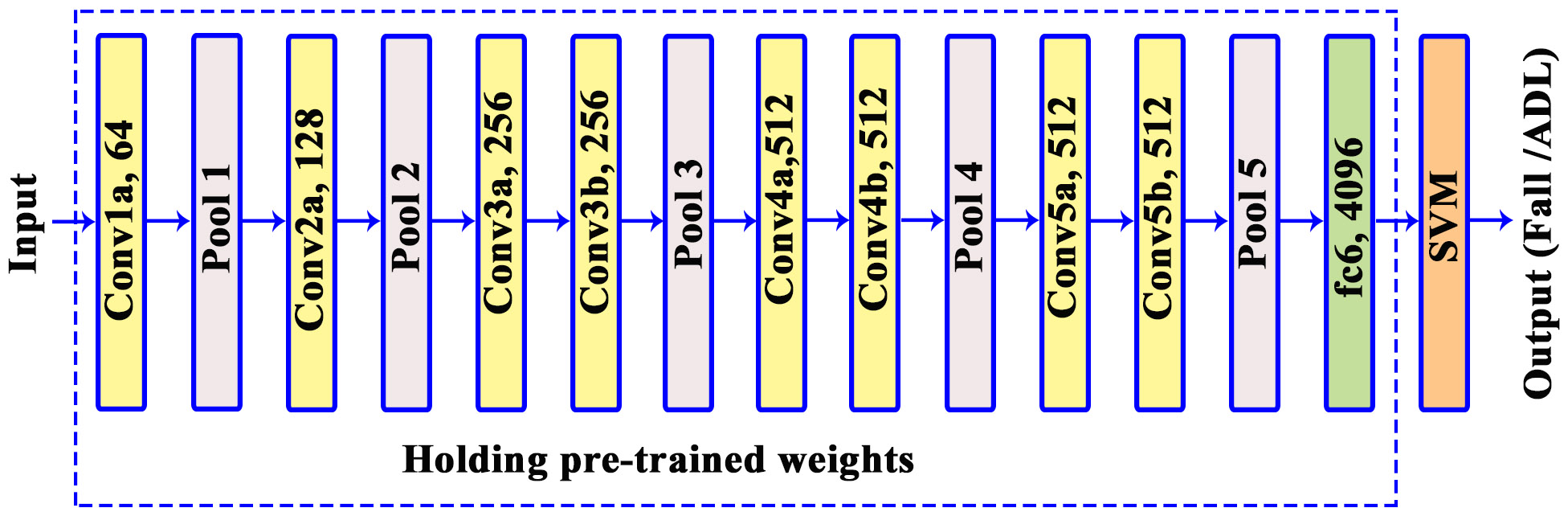}

	\caption{Proposed 3D CNN model architecture}
	\label{F_C3DArchMod}
	\centering % <-- added
\end{figure}

We used the modified version of the C3D model ``pre-trained'' on the Sport-1M dataset as a feature extractor. The Sport-1M dataset contains over one million video clips that cover a wide range of sports and physical activities, including various sports games, exercises, and outdoor activities. This dataset contains 487 classes related to sports and actions, making it suitable to use in action recognition problems like human fall detection. The architecture of the proposed model of this work is shown in Figure \ref{F_C3DArchMod}.
\begin{figure}[!htb]
	
	\centering
	
	\includegraphics[scale=0.7]{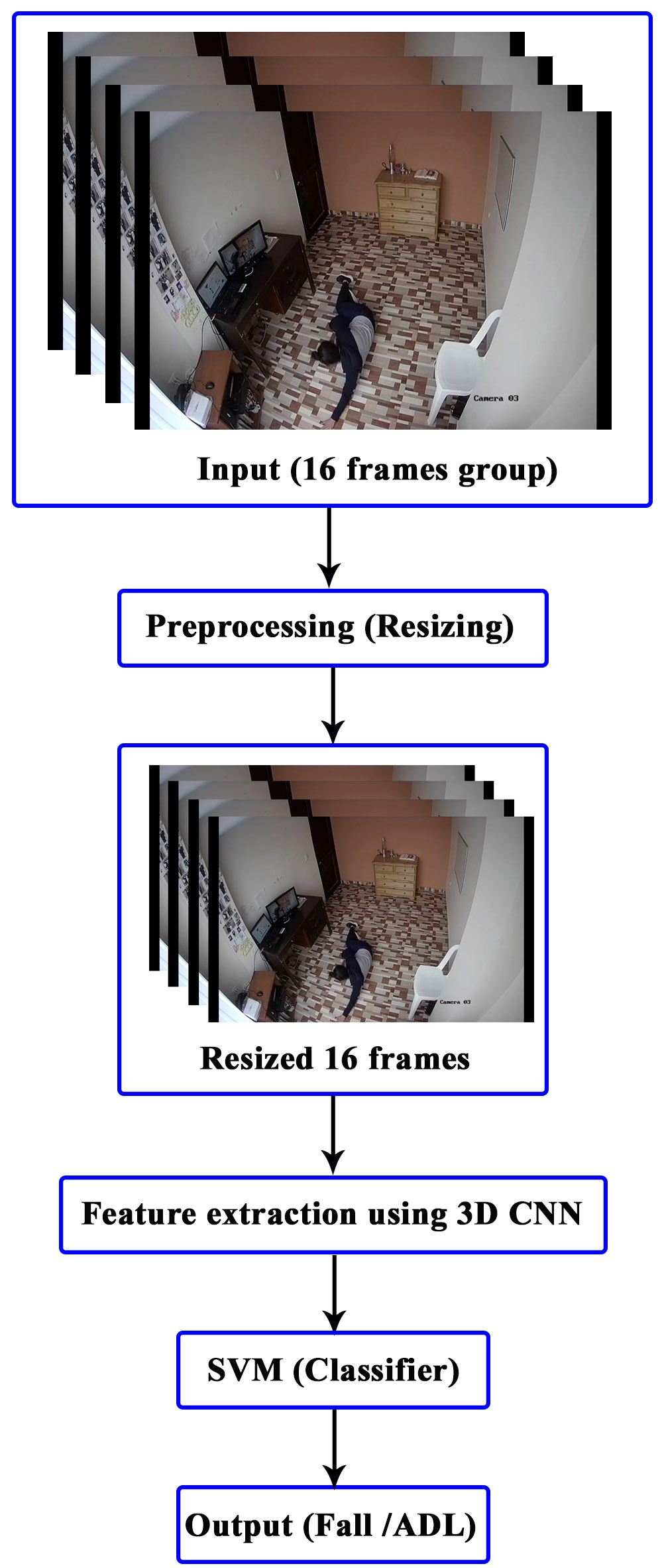}
	\vspace{-5pt}
	\caption{Proposed work methodology}
	
	\label{F_Methodology}
	
\end{figure}
This 3D CNN model comprises eight ``convolutional layers'', five ``pooling layers'', one ``fully connected'' layer \cite{alam2021leveraging}, and an SVM classifier.

The proposed methodology to detect human falls is illustrated in the flow diagram depicted in Figure \ref{F_Methodology}. At first, the input video is divided into groups (chunks) of 16 frames. These chunks are resized to 112 $\times$ 112. The resized chunks of videos are fed as the input to the proposed 3D CNN model to extract the features.
Extracted features by the proposed 3D CNN model were given as input to a linear SVM classifier to classify a frame group (chunk) as fall or ADL. The original weight of the C3D model was used to extract the features by our proposed 3D CNN model without doing any retraining. Only the SVM was trained.
We used the ``stratified shuffle split cross-validation technique'' \cite{szeghalmy2023comparative} to evaluate this work on two recent datasets (GMDCSA and CACUCAFall). We divided both datasets into five splits. For each split, 70\% of the dataset was utilized for training, while 30\% was designated for testing.

\section{Result}
\label{S_Result}

\begin{figure}[!htb]
	\centering 
	\begin{subfigure}{\textwidth}
		\includegraphics[width=.9\linewidth]{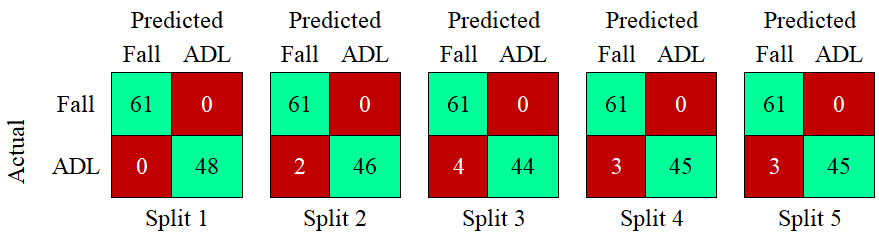}
		\caption{Confusion Matrix of the five splits of the GMDCSA Dataset}
		\label{F_CnMx_GMDCSA}
	\end{subfigure}
	
	\medskip
	
	\begin{subfigure}{\textwidth}
		\includegraphics[width=.9\linewidth]{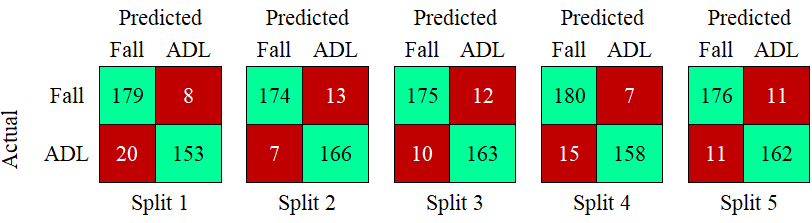}
		\caption{Confusion Matrix of the five splits of the CAUCAFall Dataset}
		\label{F_CnMx_CAUCAFall}
	\end{subfigure}
	
	\caption{Confusion Matrix of the five splits of the GMDCSA and the CAUCAFall Dataset}
	\label{F_CnMx}
\end{figure}

We evaluated the proposed model to detect human falls on two datasets, ``GMDCSA'' and ``CAUCAFall''. The ``stratified shuffle split cross-validation technique'' \cite{szeghalmy2023comparative} was used to divide the dataset into five splits. Figure \ref{F_CnMx_GMDCSA} and Figure \ref{F_CnMx_CAUCAFall} show the confusion matrices of the experiment conducted on GMDCSA and CAUCAFall datasets, respectively. The GMDCSA dataset's confusion matrices show no false negative value for any split but some false positives for each split except the first split. The false positive values for split 1, split 2, split 3, split 4, and split 5 are 0, 2, 4, 3, and 3, respectively. The GMDCSA dataset contains many sleeping activities as ADL. That is why some ADL activity was detected as fall activity.

The confusion matrices of the proposed experiment on the dataset `CAUCAFall' contain some false positive as well as false negative values because this dataset is more complex than the GMDCSA dataset. CAUCAFall dataset is developed by performing the ``fall'' and ``ADL'' activities by ten different subjects.

\begin{table}[!htbp] 
	% \scriptsize
	%	\vspace{-10pt}
	\caption{Evaluation metrics of the experiment on the dataset `GMDCSA'}
	\label{T_Res_GMDCSA}
	%	\vspace{-3pt}
	\begin{tabularx}{\linewidth}{|p{1.9cm}|X|X|X|X|X|p{2.5 cm}|} \hline
		
		\textbf{Metric} 	&\textbf{Split1}	&\textbf{Split2} 	&\textbf{Split3} 	& \textbf{Split4} 	& \textbf{Split5} 	& \textbf{Average Values}   \\ \hline
		%	File Name &Length &Description  \\ 
		Sensitivity 	&100 	&96.83 	&93.85	&95.31	&95.31	&96.26 \\ \hline
		
		Specificity 	&100	&100	&100	&100	&100	&100 \\ \hline

		Precision 		&100 	&100	&100	&100	&100	&100  \\ \hline

		Accuracy 		&100 	&98.17	&96.33	&97.25	&97.25	&97.8 \\ \hline

		F1 Score 		&100 	&98.39	&96.83	&97.60	&97.60	&98.08   \\ \hline

		FPR&0 &4.17 &8.33 & 6.25&6.25 &5 \\ \hline
		FNR &0 &0 &0 &0 &0 &0 \\ \hline
	\end{tabularx}
	%	\vspace{-10pt}
\end{table}

\begin{table}[!htb] 
	% \scriptsize
	%	\vspace{-10pt}
	\caption{Evaluation metrics of the experiment on the dataset `CAUCAFall'}
	\label{T_ResultCauca}
	%	\vspace{-3pt}
	\begin{tabularx}{\linewidth}{|p{1.9cm}|X|X|X|X|X|p{2.5cm}|} \hline
		
		\textbf{Metric} 	&\textbf{Split1}	&\textbf{Split2} 	&\textbf{Split3} 	& \textbf{Split4} 	& \textbf{Split5} 	& \textbf{Average Values}   \\ \hline
		%	File Name &Length &Description  \\ 
		Sensitivity 	&95.72 	&93.05 	&93.58	&96.26	&94.12	&94.55 \\ \hline
		
		Specificity 	&88.44	&95.95	&94.22	&91.33	&93.64	&92.72 \\ \hline

		Precision 		&89.95 	&96.13	&94.59	&92.31	&94.12	&93.42  \\ \hline

		Accuracy 		&92.22 	&94.44	&93.89	&93.89	&93.89	&93.67 \\ \hline

		F1 Score 		&92.75 	&94.57	&94.09	&94.24	&94.12	&93.95   \\ \hline

		FPR&11.56 &4.04 &5.78 &8.67 &6.35 &7.28 \\ \hline
		FNR & 4.27&6.95 &6.41 &3.74 &5.88 &5.45 \\ \hline
	\end{tabularx}
	%	\vspace{-10pt}
\end{table} 

We calculated sensitivity, specificity, precision, accuracy, F1 Score, False Positive Rate (FPR), and False Negative Rate (FNR) to measure the performance of this experiment. The values of these metrics \cite{alam2022vision} of each of the five splits are presented in Table \ref{T_Res_GMDCSA} and Table \ref{T_ResultCauca}  for the datasets GMDCSA and CAUCAFall, respectively. The average values of these metrics are also provided in the mentioned tables (Table \ref{T_Res_GMDCSA} and Table \ref{T_ResultCauca}).

\section{Conclusion and Future Scope}
\label{S_Conclusion}

In this work, we utilized transfer learning to employ the pre-trained weights of the `C3D model' up to the first ``fully connected layer'' (fc6). The extracted features were used as input for an SVM to classify the results as either a fall or an ADL. The GMDCSA and CAUCAFall datasets were employed to evaluate the proposed model. The average values of sensitivity, specificity, precision, accuracy, and F1 score for the five splits for the GMDCSA dataset were 96.26, 100, 100, 97.8, and 98.08, respectively. For the CAUCAFall dataset, the average values of sensitivity, specificity, precision, accuracy, and F1 score were 94.55, 92.72, 93.42, 93.67, and 93.95, respectively. While the results are promising, it is important to note that the datasets were generated by healthy individuals, not actual senior citizens. It would be more useful to create a new dataset containing actual fall data and test the model on this real dataset. Additionally, this work can be extended to implement an end-to-end 3D-CNN model for fall detection.

% \begin{credits}
% %\subsubsection{\ackname} A bold run-in heading in small font size at the end of the paper is
% %used for general acknowledgments, for example: This study was funded
% %by X (grant number Y).

% \subsubsection{\discintname}
% The authors have no competing interests to declare that are relevant to the content of this article.
% \end{credits}
%Author, F.: Article title. Journal \textbf{2}(5), 99--110 (2016)
%Author, F., Author, S.: Title of a proceedings paper. In: Editor,
%F., Editor, S. (eds.) CONFERENCE 2016, LNCS, vol. 9999, pp. 1--13.
%Springer, Heidelberg (2016). \doi{10.10007/1234567890}
%Author, F., Author, S., Author, T.: Book title. 2nd edn. Publisher,
%Author, A.-B.: Contribution title. In: 9th International Proceedings
%on Proceedings, pp. 1--2. Publisher, Location (2010)
%LNCS Homepage, \url{http://www.springer.com/lncs}, last accessed 2023/10/25
%\end{thebibliography}

\bibliographystyle{elsarticle-num}
\scriptsize{\bibliography{3D_CNN_TL}}
\end{document}